\documentclass[10pt,conference]{IEEEtran}

\usepackage[T1]{fontenc}
\usepackage[utf8]{inputenc}
\usepackage{lmodern}
\usepackage{amsmath,amssymb,amsfonts}
\usepackage{booktabs}
\usepackage{multirow}
\usepackage{graphicx}
\usepackage{url}
\usepackage{xcolor}
\usepackage{cite}
\usepackage{enumitem}
\usepackage{hyperref}
\graphicspath{{figures/}}
\hypersetup{colorlinks=true,citecolor=blue,linkcolor=blue,urlcolor=blue}

\title{Hybrid Visual Telemetry for Bandwidth-Constrained Robotic Vision:\\
A Pilot Study with HEVC Base Video and JPEG ROI Stills}

\author{\IEEEauthorblockN{Natalia Trukhina, Vadim Vashkelis}
\IEEEauthorblockA{Embedded Intelligence Lab \\
ntrukhina@emilab.org, vvashkelis@emilab.org}}

\begin{document}
\maketitle

\begin{abstract}
Bandwidth-constrained robotic and surveillance systems often rely on a single compressed video stream to support both continuous scene awareness and downstream machine perception. In practice, this creates a mismatch: low-bitrate video can preserve motion and coarse context, but often loses the fine local detail needed for reliable object recognition and decision-making. Motivated by a hybrid architecture in which low-resolution video supports dynamic scene understanding while event-driven high-detail regions of interest (ROIs) support close-up identification and analytics, this paper formalizes a two-channel visual telemetry scheme in which a continuous low-bitrate video stream is augmented by selectively transmitted high-detail still ROIs. This first paper does not attempt to prove the superiority of a new still-image codec. Instead, it establishes the hybrid transmission paradigm itself using a practical and reproducible codec stack: x265/HEVC for the base video stream and JPEG stills for ROI refinement. We formulate the problem as bitrate-constrained information selection for robotic vision and define an experimental protocol in which video-only and hybrid schemes are compared under matched total communication budgets. The study is designed around UAV-oriented datasets, two practical bitrate regimes, several ROI triggering policies, and object-level classification refinement on selectively transmitted ROI stills. The resulting paper lays the methodological foundation for a second-stage investigation of JPEG AI as the semantic still-image channel within the same hybrid architecture.
\end{abstract}


\section{Introduction}

Robotic platforms, UAVs, remote cameras, and mobile edge sensors frequently operate under hard bandwidth and power constraints. In such systems, one compressed video stream is often expected to serve multiple purposes simultaneously: human situational awareness, scene monitoring, object detection, classification, and decision support. Under severe compression, however, these objectives become misaligned. Temporal continuity and motion cues may remain usable, while local semantic detail degrades to the point that recognition becomes unreliable. This tension motivates the central question of this paper: under a fixed communication budget, is it better to spend all bits on one compressed video stream, or to split the budget between continuous low-rate video and sparse high-detail ROI stills?

The central idea of this paper is that bandwidth-constrained robotic vision should be treated as a selective information transport problem rather than as a pure frame-wide video compression problem. Instead of spending all available bits on uniform video coding, the communication budget can be divided into two coordinated channels: a base stream that preserves temporal continuity, and a sparse side channel that transmits high-detail information only for task-relevant objects or events. The long-term target of this research program is a hybrid architecture in which the ROI still channel may use JPEG AI. That direction is well motivated because JPEG AI has now been standardized, and its official specification explicitly targets compression and processing for both human and machine vision \cite{jpegaiworkplan}. However, the first paper should isolate the system question before the codec question.

The broader project roadmap originally targeted JPEG XL as the first still-image baseline, and JPEG XL remains relevant as a follow-on codec comparison \cite{jpegxl}. In the present pilot implementation, however, we use JPEG stills because they are robustly supported in the Colab-based experimental environment and allow us to isolate the system question before revisiting codec choice. Likewise, we use x265/HEVC for the base video stream because HEVC is a widely deployed video compression standard and x265 is a mature open-source encoder, making the experimental stack realistic and reproducible \cite{hevc_overview,x265}.

The main thesis of the paper is straightforward: for aerial robotic vision, it may be better to use part of the communication budget for continuous low-rate video and reserve the remainder for selectively transmitted high-detail object crops than to spend the entire budget on a single compressed video stream. To test this idea, we define a concrete experimental protocol on two UAV benchmarks, VisDrone and UAVDT. VisDrone provides 79 video clips with about 1.5 million annotated bounding boxes in 33{,}366 frames \cite{visdrone}, while UAVDT contains about 80{,}000 representative frames selected from 10 hours of raw UAV video \cite{uavdt}. These datasets are well aligned with the intended deployment context: small objects, aerial viewpoints, motion, scale variation, and practical surveillance-style conditions.

The contributions of this paper are fourfold. First, we formalize a hybrid visual telemetry architecture for bandwidth-constrained robotic vision. Second, we define a clean first-stage codec stack using x265/HEVC for the base stream and JPEG ROI stills. Third, we propose an evaluation design with matched total communication budgets, low and moderate bitrate regimes, and multiple ROI scheduling policies. Fourth, we establish a direct research path toward a second-stage JPEG AI study built on the same architecture and protocol.

\section{Related Work}

\subsection{Compression for Human and Machine Vision}

A growing body of work studies compression systems that jointly consider human perception and machine analysis rather than optimizing only traditional distortion metrics. Recent work explicitly frames the problem as image compression for both machine and human vision \cite{adapticmh,difficmh}. This general direction is closely aligned with the motivation of the present work, but our focus is narrower and more operational: instead of proposing a new neural codec, we study how to allocate a constrained communication budget across a continuous video channel and a sparse still-image ROI channel in robotic vision settings.

\subsection{Selective and ROI-Based Transmission}

The intuition behind ROI transmission is not new: not all regions and not all moments in a visual stream deserve equal fidelity. What distinguishes the present study is the precise system framing. The paper treats hybrid telemetry not merely as ROI coding in isolation, but as a dual-channel architecture that separates temporal continuity from semantic detail and uses event-driven ROI transport to preserve analytic value under fixed bandwidth. The first paper turns that formulation into a concrete, testable experimental protocol.

\subsection{UAV Visual Benchmarks}

VisDrone and UAVDT are especially relevant for this work because both were created to expose the difficulty of aerial-view detection and tracking. VisDrone focuses on video object detection and multi-object tracking in drone imagery and includes 79 clips and roughly 1.5 million annotations in 33{,}366 frames \cite{visdrone}. UAVDT was introduced as a large-scale UAV detection and tracking benchmark with about 80{,}000 annotated frames from 10 hours of raw videos and emphasizes challenges such as small objects, high density, and camera motion \cite{uavdt}. These characteristics make them natural testbeds for a hybrid communication approach in which low-rate video may preserve object presence while selective stills recover missing discriminative detail.

\subsection{Detection, Tracking, and Classification Backbones}

For the planned first paper, the vision backbone should be standard rather than novel. We therefore build on established models: YOLOX as a strong real-time detector, ByteTrack-style association for lightweight tracking, and ResNet-50 for ROI classification. YOLOX was introduced as a high-performance anchor-free detector \cite{yolox}, ByteTrack was proposed as a simple and effective method that associates nearly every detection box rather than discarding low-score detections \cite{bytetrack}, and ResNet provides a well-established residual architecture \cite{resnet}. More advanced detection architectures, such as the Hierarchical Instance-Conditioned Mixture-of-Experts approach \cite{himoe}, could potentially yield more robust detection under the quality-degraded conditions typical of low-bitrate video; however, using conventional backbones in this first study helps isolate the communication contribution of the paper.

\subsection{Codec Choice in the First-Stage Study}

The still-image codec choice is an important strategic issue. JPEG AI is directly relevant to the long-term vision of the project, and official JPEG and ISO material confirms that JPEG AI Part~1 is now published and explicitly designed for both human and machine vision \cite{jpegaiworkplan}. However, JPEG's own workplan also shows that profiling, reference software, conformance, and related parts continue to evolve \cite{jpegaiworkplan}. For this reason, the present preprint focuses on the system question using JPEG ROI stills in the pilot implementation and reserves JPEG XL/JPEG AI codec comparisons for a second-stage study focused specifically on codec effects within the same hybrid system.

\section{Problem Formulation}

We consider a sender that captures a full-resolution video sequence $x_1, x_2, \ldots, x_T$ and communicates with a remote receiver over a constrained link. A conventional system uses a single compressed video stream $v_{1:T}$ and assigns the entire communication budget to that stream. In the proposed hybrid system, the communication budget is divided into two channels:
\begin{enumerate}[leftmargin=1.5em]
    \item a continuous low-bitrate base video stream,
    \item a sparse side channel carrying high-detail still images of selected ROIs.
\end{enumerate}

Let $B_{\text{tot}}$ denote the total average communication budget over a time window. Let $B_v$ be the average bitrate allocated to the base stream and $B_r$ the average bitrate allocated to ROI stills, with the budget constraint
\begin{equation}
B_v + B_r \leq B_{\text{tot}}.
\end{equation}

At time $t$, the system observes a set of candidate ROIs $R_t = \{r_t^1, r_t^2, \ldots, r_t^{K_t}\}$, derived from detections and tracks in the decoded base video. For each candidate, the scheduler decides whether to transmit a still-image crop extracted from the original source frame.

The goal of the system is not to minimize frame-wise distortion but to maximize downstream robotic vision utility. In this first paper, the relevant utility combines two tasks: object detection continuity from the base video stream and object classification refinement from selectively transmitted ROI stills. We therefore define the objective as
\begin{equation}
\max_{\pi,\, B_v,\, B_r} \; \mathbb{E}\left[U_{\text{det}} + \lambda U_{\text{cls}}\right]
\quad \text{s.t.} \quad
B_v + B_r \leq B_{\text{tot}},
\end{equation}
where $\pi$ is the ROI scheduling policy, $U_{\text{det}}$ denotes detection utility, $U_{\text{cls}}$ denotes classification utility, and $\lambda$ controls the relative weight of classification refinement.

\section{Proposed Hybrid Visual Telemetry System}

\subsection{System Overview}

\begin{figure*}[t]
\centering
\includegraphics[width=\textwidth]{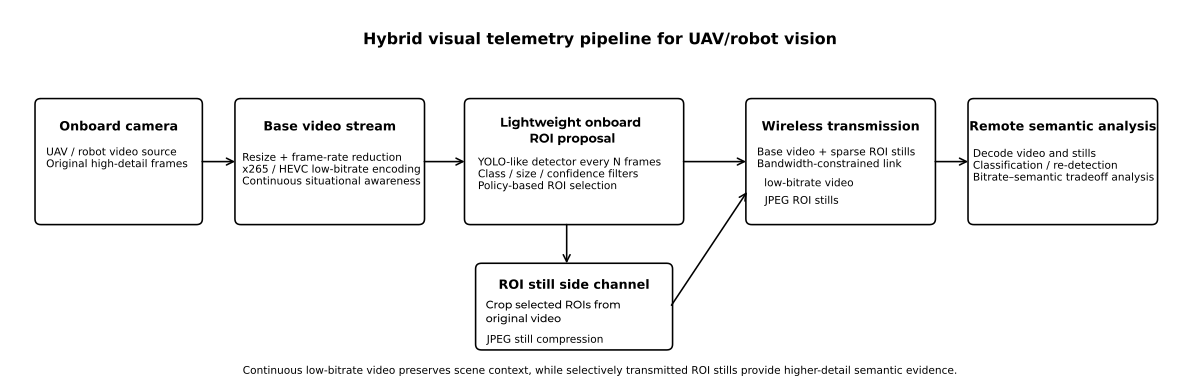}
\caption{Hybrid visual telemetry pipeline for bandwidth-constrained robotic vision. Onboard, the full-resolution camera feed is split into a continuous low-bitrate HEVC base stream and a sparse ROI still side channel. A lightweight onboard detector proposes candidate regions, and a policy-based selector determines which ROIs are transmitted as JPEG stills. Both channels traverse the bandwidth-constrained link and are decoded at the remote side for combined detection, tracking, and classification refinement.}
\label{fig:pipeline}
\end{figure*}

Fig.~\ref{fig:pipeline} illustrates the overall processing pipeline of the proposed hybrid visual telemetry system. The sender processes each input frame using two branches. The first branch encodes the full frame into a low-bitrate HEVC stream using x265. This branch is intended to preserve temporal continuity, motion, context, and object presence. The second branch generates ROI candidates from the visual content and, when the scheduler decides to do so, extracts object crops directly from the original source frames and encodes them as JPEG still images. These ROI stills are then sent together with lightweight metadata such as frame index, bounding-box coordinates, and track identity.

At the receiver, the base stream is decoded continuously and fed into detection and tracking. When an ROI still arrives, it is classified independently and used to refine the class prediction of the corresponding tracked object.

\subsection{Detection and Tracking}

The present study is not about inventing a new detector or tracker, so we intentionally choose standard backbones. YOLOX provides the detector, and ByteTrack-style association is used for linking detections into tracks \cite{yolox,bytetrack}. For every decoded video frame, the detector produces object boxes and confidence scores. The tracker maintains short object trajectories and supports object persistence across frames. Each active track becomes a candidate for ROI refinement. We note that more specialized detection architectures, such as instance-conditioned Mixture-of-Experts models \cite{himoe}, could replace YOLOX in future work to improve detection robustness on compressed video, but this is orthogonal to the communication question addressed here.

\subsection{ROI Extraction and Still-Image Encoding}

\begin{figure}[t]
\centering
\includegraphics[width=\columnwidth]{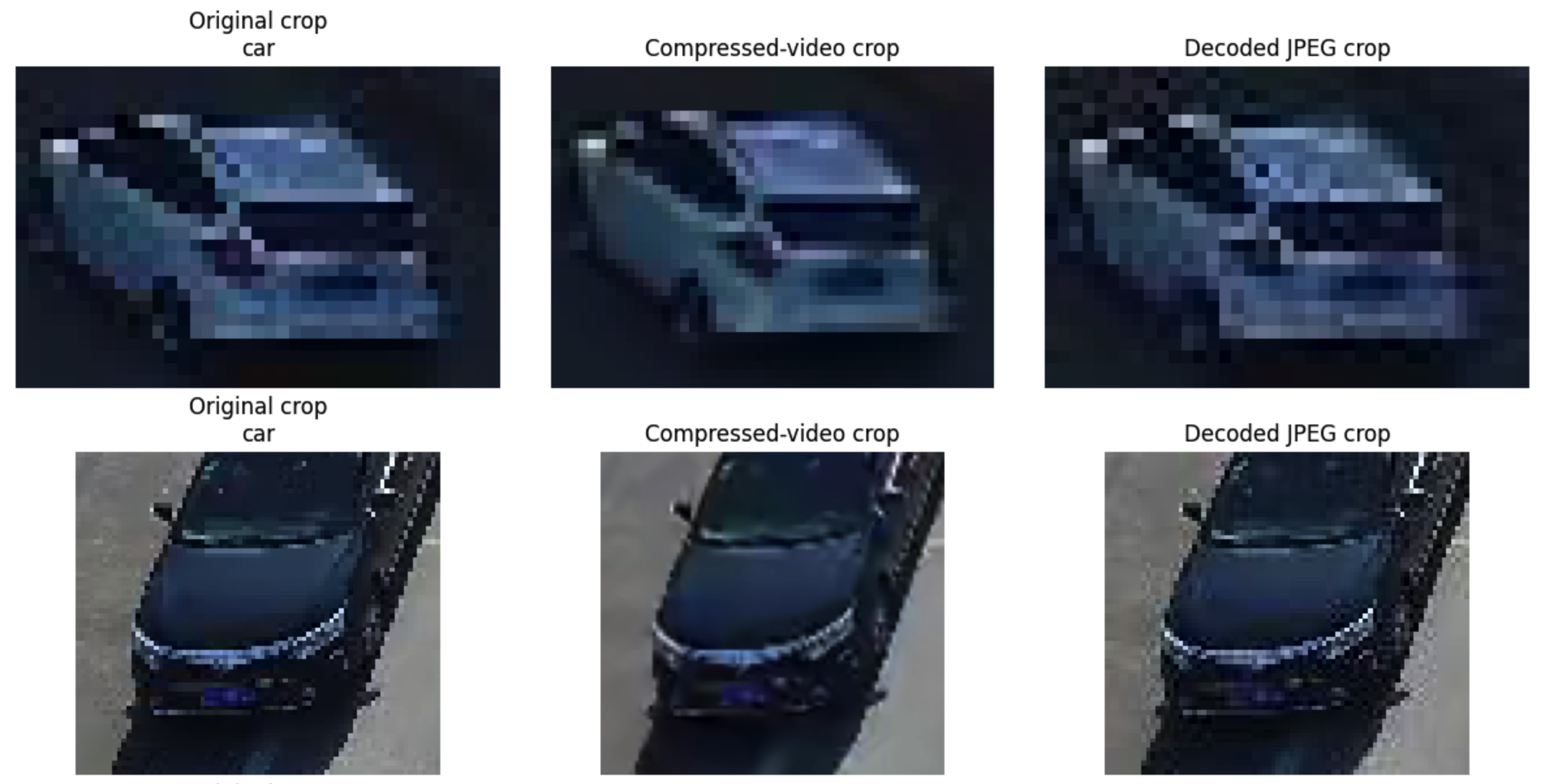}\\\smallskip
\includegraphics[width=\columnwidth]{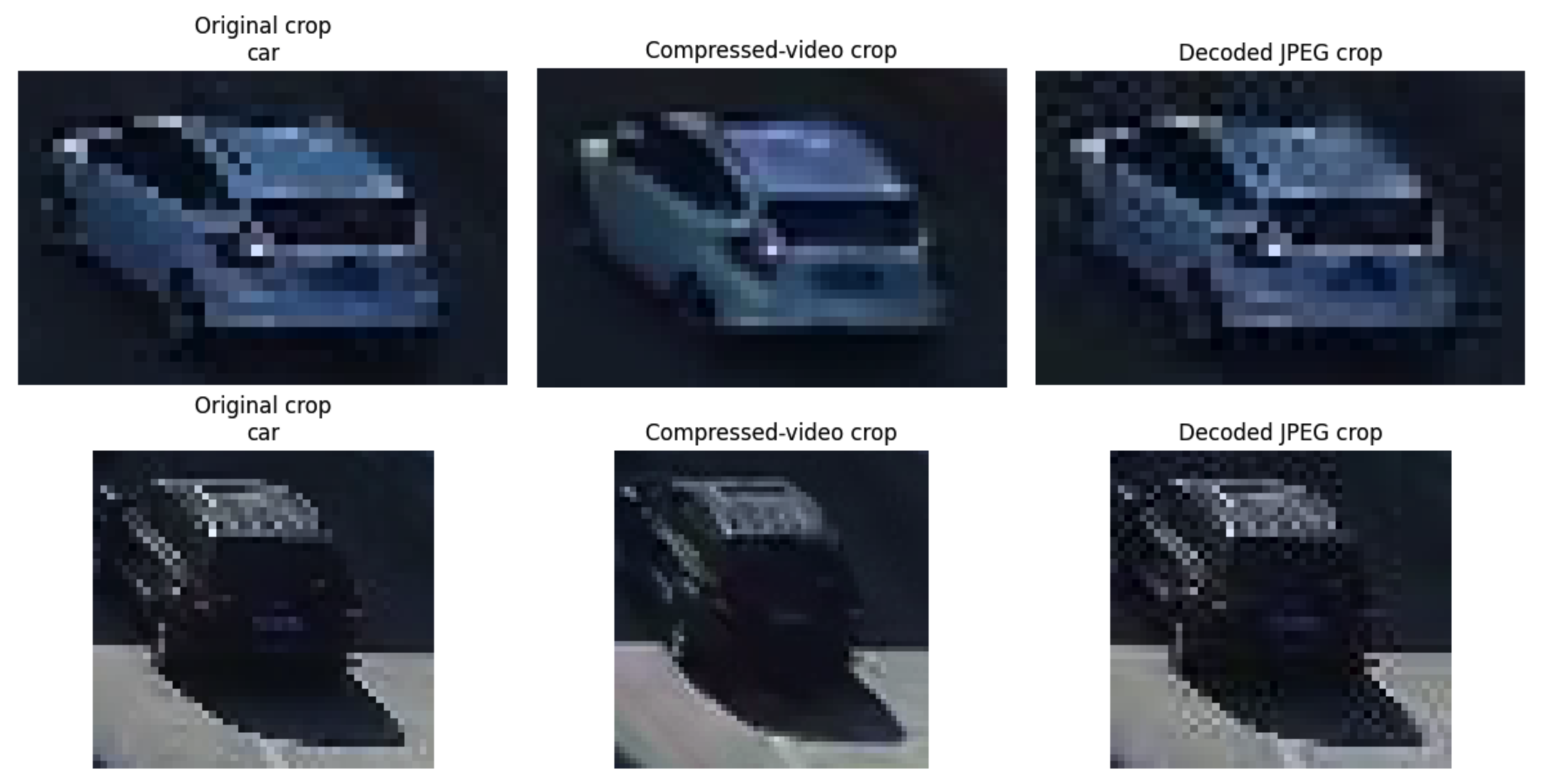}
\caption{Visual comparison of object crops under three conditions for two representative vehicles from the pilot UAVDT sequence. In each row: (left) original crop from the uncompressed source frame, (center) crop from the decoded low-bitrate HEVC stream, and (right) crop from the decoded JPEG still. The compressed-video crops exhibit blocking artifacts and loss of fine detail, while the JPEG still crops retain substantially more of the original structure, supporting more reliable classification.}
\label{fig:crop_comparison}
\end{figure}

ROI crops are extracted from the original uncompressed source frames rather than from the decoded low-bitrate video. This design is critical: the purpose of the side channel is to recover information that was never preserved in the compressed base stream. Fig.~\ref{fig:crop_comparison} illustrates this point with representative vehicle crops from the pilot UAVDT sequence. The original crops retain fine edges and local texture; the compressed-video crops suffer from blocking artifacts and motion-related blur that obscure discriminative features; the decoded JPEG still crops, although also lossy, preserve substantially more structural detail from the original source. Each ROI is formed by taking the tracked object bounding box plus a fixed padding margin to retain local context. The crop is then resized appropriately for the classifier input and encoded as a JPEG still for transmission.

In the original project roadmap we intended to start with JPEG XL and later transition to JPEG AI. In the present pilot implementation, however, we use JPEG because it is robustly available in the experimental Colab environment and allows us to isolate the system question before revisiting the codec question. The second-stage study will return to JPEG XL and JPEG AI once the selector and evaluation protocol are fully stabilized.

\subsection{Classification Refinement}

Classification is performed using ResNet-50 \cite{resnet}. In the video-only baseline, the classifier operates on object crops taken from the decoded video stream. In the hybrid setting, the same classifier operates on decoded JPEG ROI stills when they are available. This ensures that the comparison isolates the effect of communication strategy rather than the classifier architecture.

Whenever an ROI-based classification is available for an active track, that prediction replaces the current class estimate derived from the decoded base-video crop. More elaborate fusion strategies are possible, but the replacement rule is appropriate for the first paper because it makes the contribution of the ROI channel directly observable.

\section{ROI Scheduling Policies}

The existence of a still-image side channel is not sufficient on its own; the system must also decide when that channel should be used. We therefore compare five transmission policies.

\textbf{M0: Video-only.} The full budget is assigned to x265 video. No ROI stills are transmitted.

\textbf{M1: Periodic hybrid.} ROI stills are sent for active tracks at a fixed interval. This serves as a naïve hybrid baseline.

\textbf{M2: First-appearance hybrid.} An ROI still is sent when a new track first appears. This models sparse event-driven transmission with minimal repeated updates.

\textbf{M3: Uncertainty-triggered hybrid.} An ROI still is sent when the detector confidence or classifier confidence from the decoded video is below a threshold.

\textbf{M4: Size-triggered hybrid.} An ROI still is sent when the detected object is small in the decoded frame, since small aerial objects are especially vulnerable to compression loss.

\textbf{M5: Utility-per-bit hybrid.} This is the proposed policy. Each candidate ROI is assigned a normalized score
\begin{equation}
S(r)=\frac{0.5\,U(r)+0.3\,S_{\text{small}}(r)+0.2\,N(r)}{C(r)},
\end{equation}
where $U(r)$ is an uncertainty term, $S_{\text{small}}(r)$ is a size-priority term, $N(r)$ is a novelty term indicating whether the object is new or has not been recently refined, and $C(r)$ is an estimate of the ROI transmission cost in bits. At each decision step, the system selects the highest-scoring ROI whose score exceeds a threshold and whose transmission does not violate the rolling ROI bitrate budget.

\section{Experimental Design}

\subsection{Datasets}

The experiments are designed around two UAV datasets.

\textbf{VisDrone} is used as the primary benchmark. According to the ECCV 2018 challenge paper, the VisDrone video benchmark contains 79 video clips and about 1.5 million annotated bounding boxes in 33{,}366 frames, with tasks including video object detection and multi-object tracking \cite{visdrone}.

\textbf{UAVDT} is used as the secondary benchmark. The original ECCV 2018 paper describes UAVDT as a large-scale UAV detection and tracking benchmark containing about 80{,}000 representative frames selected from 10 hours of raw UAV videos \cite{uavdt}.

These datasets were selected because they reflect the intended deployment context of the paper: aerial viewpoints, scale variation, small objects, camera motion, and practical surveillance-like scenes.

\subsection{Resolution and Frame Rate}

All experiments are defined at $1280\times720$ resolution and 15 frames per second. This operating point is representative of realistic UAV and robotic telemetry and places the experiments in a regime where bandwidth remains constrained but the stream is still practically useful.

\subsection{Communication Regimes}

The paper defines two operating regimes, but the current preprint reports a pilot evaluation in the \textbf{low-bitrate regime}. In this regime the target base-video bitrate is 0.8 Mbps and the measured achieved bitrate in the reported sequence is 0.801 Mbps. The moderate-bitrate regime remains part of the protocol and will be reported in a later revision.

\subsection{Codec Stack}

The base stream is encoded with x265/HEVC \cite{hevc_overview,x265}. The ROI side channel in the present pilot uses JPEG still images. This is an implementation convenience rather than the long-term codec choice; JPEG XL and JPEG AI remain part of the follow-on research plan.

\subsection{Metrics}

The experiments are organized around matched total communication budgets, but the current preprint emphasizes \emph{proxy semantic-gain metrics} rather than final task accuracy. The primary reported metrics are total communicated bitrate, number of selected ROIs, ROI bitrate share, mean ROI payload size, mean confidence gain between decoded video crops and decoded still crops, positive confidence-gain rate, prediction-change rate, and entropy reduction. We also report small-object subsets because hybrid refinement is expected to be most beneficial for small aerial objects.

\subsection{Comparison Protocol}

For each dataset and bitrate regime, we first construct the video-only baseline by encoding the source video with x265 at the full target bitrate and running detection, tracking, and classification on the decoded stream.

We then construct hybrid variants by reserving part of the communication budget for JPEG ROI stills and transmitting those stills only for selected frames and selected detections according to the scheduling policy. ROI stills are always cropped from the original source video rather than the compressed base stream.

All comparisons are made at matched total communication budgets:
\begin{equation}
B_{\text{tot}} = B_v + B_r.
\end{equation}
This is the key fairness constraint of the paper. If the hybrid system outperforms video-only transmission, it should do so because it uses the same budget more intelligently, not because it transmits more total information.

\section{Method and Experiment Matrix}

Table~\ref{tab:methods} summarizes the method matrix, and Table~\ref{tab:metrics} summarizes the evaluation metrics.

\begin{table*}[t]
\centering
\small
\caption{Method matrix for the first-stage study. All hybrid methods use x265/HEVC for the base stream and JPEG ROI stills. Bitrate pairs show base-video/ROI-still allocation in Mbps. The current preprint reports pilot results for the selector-sweep subset discussed in the text.}
\label{tab:methods}
\begin{tabular}{llccp{2.8cm}p{2.5cm}}
\toprule
ID & Scheme & Low (V/R) & Mod. (V/R) & Trigger Rule & Purpose \\
\midrule
M0 & Video-only & 0.80/0.00 & 1.40/0.00 & No ROI transmissions & Main baseline \\
M1 & Periodic hybrid & 0.65/0.15 & 1.20/0.20 & One ROI every 15 frames for active tracks & Naïve hybrid baseline \\
M2 & First-appearance & 0.65/0.15 & 1.20/0.20 & ROI at track creation or re-init. & Sparse event-driven baseline \\
M3 & Uncertainty-trig. & 0.65/0.15 & 1.20/0.20 & ROI when det./cls.\ confidence below threshold & Confidence repair baseline \\
M4 & Size-triggered & 0.65/0.15 & 1.20/0.20 & ROI when object area below threshold & Small-object recovery baseline \\
M5 & Utility-per-bit & 0.65/0.15 & 1.20/0.20 & Highest-scoring ROI under budget (Eq.~3) & Proposed method \\
\bottomrule
\end{tabular}
\end{table*}

\begin{table}[t]
\centering
\small
\caption{Evaluation metrics reported in the planned experiments.}
\label{tab:metrics}
\begin{tabular}{p{2.5cm}p{4.2cm}}
\toprule
Metric & Role \\
\midrule
Detection mAP & Localization quality from base stream \\
Cls.\ Top-1 Acc. & Object-level recognition \\
Macro-F1 & Class-imbalance robustness \\
Total bitrate & Primary fairness axis \\
ROI stills/min & Side-channel activity \\
Mean ROI bytes & Still-codec efficiency proxy \\
Tracks refined & ROI refinement coverage \\
End-to-end latency & Deployment relevance \\
Acc.\ by obj.\ size & Small-object gain diagnostics \\
\bottomrule
\end{tabular}
\end{table}

\section{Pilot Results}

\subsection{Pilot Setup}

The current preprint reports a pilot end-to-end evaluation on a single UAVDT sequence in the low-bitrate regime. The base video was encoded with x265 at a measured achieved bitrate of 0.801 Mbps. ROI candidates were generated by a lightweight YOLO detector operating on the encoded stream every fifth frame over a 60-frame processed subset. The ROI side channel used JPEG stills cropped from the original video, and semantic comparison was performed with an ImageNet-pretrained ResNet-50 applied to (i) decoded compressed-video crops and (ii) decoded JPEG still crops.

Because this is an initial proof-of-concept study, we report proxy semantic metrics rather than final task accuracy against trusted class labels. In particular, we evaluate confidence gain, entropy reduction, and prediction-change statistics between the video-crop branch and the still-crop branch.

\subsection{Selector Sweep}

Before running the full end-to-end chain, we applied several selector policies to the same set of raw detections. Table~\ref{tab:selector_sweep} summarizes the resulting sparsity levels. The key observation is that policy design has a first-order effect on the cost of the side channel. In the pilot sequence, the permissive selector kept 178 ROI candidates from 325 raw detections, whereas the strict and balanced selectors reduced this count to the range of 57--110 while still preserving nearly complete frame-level coverage.

\begin{table}[t]
\centering
\small
\caption{Selector sweep on one low-bitrate UAVDT pilot sequence. Numbers are computed before still compression and semantic evaluation.}
\label{tab:selector_sweep}
\begin{tabular}{p{2.2cm}ccc}
\toprule
Policy & Sel.\ ROIs & Sel.\ Ratio & Frame Cov. \\
\midrule
Permissive & 178 & 0.548 & 1.000 \\
conf\_size\_top1 & 59 & 0.182 & 0.983 \\
strict\_small\_only & 57 & 0.175 & 0.950 \\
balanced\_top2 & 110 & 0.338 & 0.983 \\
\bottomrule
\end{tabular}
\end{table}

These selector-only results already show why a policy sweep is necessary: the hybrid system can behave either like an almost always-on auxiliary stream or like a genuinely sparse side channel depending on the trigger and ranking logic.

\subsection{End-to-End Hybrid Results}

We ran the full end-to-end pipeline for two representative selector policies: a strict low-cost policy (\texttt{conf\_size\_top1}) and a more balanced policy (\texttt{balanced\_top2}). Table~\ref{tab:pilot_results} summarizes the resulting bitrate and semantic metrics.

\begin{table*}[t]
\centering
\footnotesize
\caption{End-to-end pilot results for two policy settings on one low-bitrate UAVDT sequence. Positive confidence gain and positive entropy gain indicate that the still-image branch is helping on average.}
\label{tab:pilot_results}
\setlength{\tabcolsep}{3.5pt}
\begin{tabular}{lcccccccccc}
\toprule
Policy & ROIs & Rate (Hz) & Br. (Mbps) & Share & Bytes & Vid.\ Conf. & Still Conf. & $\Delta$Conf. & Pos.\ Rate & $\Delta$Entropy \\
\midrule
conf\_size\_top1 & 59 & 1.123 & 0.0123 & 0.0151 & 1364 & 0.159 & 0.154 & $-$0.005 & 0.542 & 0.097 \\
balanced\_top2 & 110 & 2.094 & 0.0258 & 0.0312 & 1540 & 0.149 & 0.166 & $+$0.016 & 0.600 & 0.209 \\
\bottomrule
\end{tabular}
\end{table*}

Several points are worth emphasizing.

First, the strict selector dramatically reduces side-channel cost. For \texttt{conf\_size\_top1}, the ROI bitrate is only 0.0123 Mbps, corresponding to 1.5\% of the total hybrid bitrate. This is a very attractive operating point from the deployment perspective. However, the same selector is somewhat too aggressive semantically: the mean confidence gain becomes slightly negative ($-0.0047$), even though 54.2\% of selected ROIs still yield positive confidence gain and the mean entropy gain remains positive.

Second, the balanced selector restores positive average semantic benefit while keeping the side channel small. For \texttt{balanced\_top2}, the ROI bitrate is still only 0.0258 Mbps, or 3.1\% of the total hybrid bitrate, yet the mean confidence gain rises to $+0.0164$, 60.0\% of selected ROIs yield positive confidence gain, and the mean entropy gain rises to 0.2087. This is the strongest operating point observed in the present pilot study.

Third, the policy sweep reveals a clear bitrate--utility tradeoff. The strict selector is a strong efficiency extreme, whereas the balanced selector offers a better compromise between ROI sparsity and semantic benefit. In other words, the pilot experiments support the central systems hypothesis of the paper: selective ROI transmission can provide measurable semantic refinement at a modest communication cost, but the selector must be tuned carefully.

\subsection{Interpretation}

The current results should be read as a proof-of-concept rather than as a final benchmark claim. The classifier is out-of-domain for aerial traffic imagery, the labels are proxy predictions rather than trusted task labels, and the evaluation is presently limited to a single sequence and the low-bitrate regime. Nevertheless, the measured tradeoff is already informative. The balanced policy achieves positive mean confidence gain while keeping ROI-side traffic extremely small in absolute terms. The strict policy demonstrates that the side channel can be pushed down to near-minimal overhead, although at some semantic cost. Together, these two points define a useful operating range for subsequent work. Qualitatively, the visual evidence in Fig.~\ref{fig:crop_comparison} confirms that the still-image side channel preserves discriminative detail that the compressed base stream loses, which is the prerequisite for the observed confidence and entropy gains.

\section{Discussion}

The present preprint should be understood as an initial systems study. Its main contribution is not a finalized codec comparison, but the demonstration that a low-bitrate video stream plus selectively transmitted still ROIs can be implemented end to end, instrumented with bitrate accounting, and evaluated with policy-sensitive semantic metrics. The experiments already show that selector design is crucial: an overly permissive policy makes the side channel too dense, while an overly strict policy can erase the average semantic benefit. A balanced policy appears to offer the most promising compromise in the current setup.

The current implementation also reveals the next technical priorities. First, frame-level coverage remains high even for the stricter policies, which suggests that temporal suppression, cooldown logic, or lightweight tracking should be added so that the system does not send nearly one ROI on every processed frame. Second, the semantic probe is intentionally generic. A stronger detector/classifier stack or a manually verified subset will make later versions of the paper more definitive. Third, the still-image codec in the pilot implementation is JPEG for practical reasons. This does not weaken the systems contribution of the present paper, but it does leave open the codec-level comparison with JPEG XL and JPEG AI that motivated the broader research program.

In that sense, the current results are useful precisely because they isolate the system question. They show that hybrid selective transmission is promising before the project invests in more specialized models, broader multi-sequence evaluation, or advanced still-image codecs.

\section{Conclusion}

This paper presented an initial end-to-end study of hybrid visual telemetry for bandwidth-constrained robotic vision. The proposed architecture combines a continuous low-bitrate HEVC stream with selectively transmitted still-image ROIs cropped from the original video. On a pilot UAVDT sequence in the low-bitrate regime, the experiments showed that ROI policy design strongly controls the tradeoff between communication cost and semantic benefit. A strict selector reduced ROI-side traffic to only 1.5\% of total hybrid bitrate but became slightly too aggressive semantically, whereas a balanced selector retained positive mean confidence gain while keeping ROI-side traffic to only 3.1\% of the total bitrate.

These findings are sufficient to support the central claim of the paper: hybrid video-plus-still transmission is a viable and promising approach for low-bandwidth robotic vision, provided that ROI selection is treated as a bitrate-aware policy problem rather than a simple detection pass-through. The present preprint therefore serves as a proof-of-concept foundation for broader multi-sequence evaluation, moderate-bitrate experiments, temporal suppression policies, and future codec-level comparisons involving JPEG XL and JPEG AI.

\section*{Acknowledgment}
The source code for reproducing the experiments reported in this paper is publicly available at \url{https://gitlab.com/emilab-group/hybrid-robotic-vision-for-uav}. This manuscript reports a pilot proof-of-concept study and is intended as a foundation for subsequent expanded experiments and revisions.

\end{document}